%% file: main.tex
\documentclass[letterpaper, 10 pt, journal, twoside]{ieeetran}

\IEEEoverridecommandlockouts                              

\input{preamble.tex}

\usepackage{caption}
\newcommand{\citep}[1]{\cite{#1}}
\usepackage{authblk}

\title{
Low Level Control of a Quadrotor with \\ Deep Model-Based Reinforcement Learning
}

\author{Nathan O. Lambert\textsuperscript{1}, Daniel S. Drew\textsuperscript{1}, Joseph Yaconelli\textsuperscript{2},\\ Sergey Levine\textsuperscript{1}, Roberto Calandra\textsuperscript{3}, and Kristofer S. J. Pister\textsuperscript{1}
\thanks{Manuscript received: February, 24, 2019; Revised June, 3, 2019; Accepted July, 7, 2019.}
\thanks{This paper was recommended for publication by Editor Tamim Asfour upon evaluation of the Associate Editor and Reviewers' comments. }
\thanks{Corresponding author: Nathan O. Lambert \href{mailto:nol@berkeley.edu}{nol@berkeley.edu}}%
\thanks{\textsuperscript{1}Department of Electrical Engineering and Computer Sciences, University of California, Berkeley.}
\thanks{\textsuperscript{2}Author is supported by the Berkeley SUPERB REU Program.}
\thanks{\textsuperscript{3}Facebook AI Research, Menlo Park, CA}%
\thanks{Digital Object Identifier (DOI): see top of this page.}
}

\begin{document}

\maketitle



\begin{abstract}
\input{0_abstract.tex}

\end{abstract}



\input{1_introduction.tex}

\input{2_relatedwork.tex}
\input{3_groundwork.tex}

\input{4_modeltraining.tex}
\input{5_control.tex}

\input{6_eval.tex}
\input{7_discussion.tex}

\input{8_conclusion.tex}


\section*{ACKNOWLEDGMENT}
The authors thank the UC Berkeley Sensor~\&~Actuator Center (BSAC), Berkeley DeepDrive, and  Nvidia Inc.

\bibliographystyle{IEEEtran}
\bibliography{mbrl}

\input{_appendices.tex}

\end{document}

%% file: preamble.tex
\usepackage[top=.794in, bottom=.599in, left=.668in, right=.668in]{geometry}

\usepackage{graphicx} 
\usepackage{amsmath} 

\DeclareMathOperator*{\argmin}{arg\,min}

\usepackage{csvsimple}  

\usepackage{tikz}
\usetikzlibrary{positioning, shapes, arrows.meta}
\usetikzlibrary{shapes,arrows,chains}
\usetikzlibrary{arrows,calc,automata}

\usepackage{afterpage}
\usepackage{placeins}

\usepackage{verbatim}
\usepackage{hyperref}
\usepackage{nameref}
\usepackage{algorithm}
\usepackage{algorithmic}
\usepackage{gensymb} 
\usepackage{siunitx}

\tikzset{
    block/.style = {draw, rectangle, 
        minimum height=.5cm, 
        minimum width=1cm},
    input/.style = {coordinate,node distance=.5cm},
    output/.style = {coordinate,node distance=.5cm},
    arrow/.style={draw, -latex,node distance=.5cm},
    pinstyle/.style = {pin edge={latex-, black,node distance=1cm}},
    sum/.style = {draw, circle, node distance=1cm},
    line/.style={-{Stealth}}
    }

\usepackage{cite}

\usepackage{amsmath,amssymb}

\usepackage{array}

\usepackage{upgreek}
\usepackage{float}          
\usepackage{stfloats}

\usepackage{float}
\floatstyle{plain}
\restylefloat{table}
\usepackage{graphicx}

\renewcommand{\sec}[1]{Section~\ref{#1}}
\newcommand{\fig}[1]{Figure~\ref{#1}}
\newcommand{\eq}[1]{Equation~\eqref{#1}}

%% file: 0_abstract.tex
Designing effective low-level robot controllers often entail platform-specific implementations that require manual heuristic parameter tuning,  significant system knowledge, or long design times.
With the rising number of robotic and mechatronic systems deployed across areas ranging from industrial automation to intelligent toys, the need for a general approach to generating low-level controllers is increasing.
To address the challenge of rapidly generating low-level controllers, we argue for using model-based reinforcement learning (MBRL) trained on relatively small amounts of automatically generated (i.e., without system simulation) data.
In this paper, we explore the capabilities of MBRL on a Crazyflie centimeter-scale quadrotor with rapid dynamics to predict and control at $\leq$ 50Hz.
To our knowledge, this is the first use of MBRL for controlled hover of a quadrotor using only on-board sensors, direct motor input signals, and no initial dynamics knowledge. 
Our controller leverages rapid simulation of a neural network forward dynamics model on a GPU-enabled base station, which then transmits the best current action to the quadrotor firmware via radio.
In our experiments, the quadrotor achieved hovering capability of up to 6 seconds with 3 minutes of experimental training data.

%% file: 1_introduction.tex
\section{Introduction}

\IEEEPARstart{T}{he} ideal method for generating a robot controller would be extremely data efficient, free of requirements on domain knowledge, and safe to run. 
Current strategies to derive low-level controllers are effective across many platforms, but system identification often requires substantial setup and experiment time while PID tuning requires some domain knowledge and still results in dangerous roll-outs. 
With the goal to reduce reliance on expert-based controller design, in this paper we investigate the question:
Is it possible to autonomously learn competitive low-level controllers for a robot, without simulation or demonstration, in a limited amount of time?
\begin{figure}[t]
    \centering
    \includegraphics[width=.9\columnwidth]{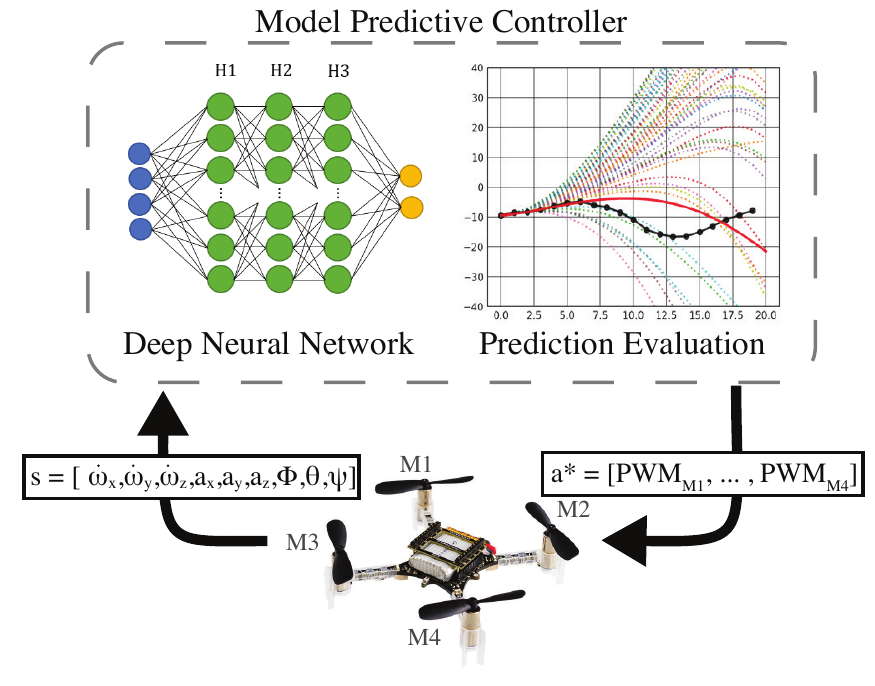}
    \caption{The model predictive control loop used to stabilize the Crazyflie. 
    Using deep model-based reinforcement learning, the quadrotor reaches stable hovering with only 10,000 trained datapoints -- equivalent to 3 minutes of flight.}
    \label{fig:sysblock}
\end{figure}

To answer this question, we turn to model-based reinforcement learning~(MBRL) -- a compelling approach to synthesize controllers even for systems without analytic dynamics models and with high cost per experiment~\citep{gpcontrol}. 
MBRL has been shown to operate in a data-efficient manner to control robotic systems by iteratively learning a dynamics model and subsequently leveraging it to design controllers~\citep{Williams2017}.
Our contribution builds on simulated results of MBRL~\citep{Chua2018}.
We employ the quadrotor as a testing platform to broadly investigate controller generation on a highly nonlinear, challenging system, not to directly compare performance versus existing controllers.
This paper is the first demonstration of controlling a quadrotor with direct motor assignments sent from a MBRL derived controller learning only via experience.
Our work differs from recent progress in MBRL with quadrotors by exclusively using experimental data and focusing on low level control, while related applications of learning with quadrotors employ low-level control generated in simulation~\cite{Hwangbo2017} or use a dynamics model learned via experience to command on-board controllers~\cite{Bansal2016}. 
Our MBRL solution, outlined in~\fig{fig:sysblock},  employs neural networks (NN) to learn a forwards dynamics model coupled with a `random shooter' MPC, which can be efficiently parallelized on a graphic processing unit~(GPU) to execute low-level, real-time control.

Using MBRL, we demonstrate controlled hover of a Crazyflie via on-board sensor measurements and application of  pulse width modulation~(PWM) motor voltage signals.
Our method for quickly learning controllers from real-world data is not yet an alternative to traditional controllers such as PID, but it opens important avenues of research.
The general mapping of the forward dynamics model, in theory, allows the model to be used for control tasks beyond attitude control.
Additionally, we highlight the capability of leveraging the predictive models learned on extremely little data for working at frequencies $\leq$\SI{50}{\hertz}, while a hand tuned PID controller at this frequency failed to hover the Crazyflie. 
With the benefits outlined, the current MBRL approach has limitations in performance and applicability to our goal of use with other robots. 
The performance in this paper has notable room for improvement by mitigating drift.
Future applications are limited by our approach's requirement of a high power external GPU -- a prohibitively large computational footprint when compared to standard low-level controllers -- and by the method's potential for collisions when learning.

The resulting system achieves repeated stable hover of up to 6 seconds, with failures due to drift of unobserved states, within 3 minutes of fully-autonomous training data.
These results demonstrate the ability of MBRL to control robotic systems in the absence of \textit{a priori} knowledge of dynamics, pre-configured internal controllers for stability or actuator response smoothing, and expert demonstration.

%% file: 2_relatedwork.tex
\section{Related Work}

\subsection{Attitude and Hover Control of Quadrotors}
Classical controllers (e.g., PID, LQR, iLQR) in conjunction with analytic models for the rigid body dynamics of a quadrotor are often sufficient to control vehicle attitude~\cite{mahony2012multirotor}. 
In addition, linearized models are sufficient to simultaneously control for global trajectory attitude setpoints using well-tuned nested PID controllers~\cite{mellinger2012trajectory}. 
Standard control approaches show impressive acrobatic performance with quadrotors, but we note that we are not interested in comparing our approach to finely-tuned performance; the goal of using MBRL in this context is to highlight a solution that automatically generates a functional controller in less or equal time than initial PID hand-tuning, with no foundation of dynamics knowledge.

Research focusing on developing novel low-level attitude controllers shows functionality in extreme nonlinear cases, such as for quadrotors with a missing propeller ~\cite{zhang2016controllable}, with multiple damaged propellers~\cite{mueller2014stability}, or with the capability to dynamically tilt its propellers~\cite{ryll2012modeling}. Optimal control schemes have demonstrated results on standard quadrotors with extreme precision and robustness \cite{liu2016robust}. 

Our work differs by specifically demonstrating the possibility of attitude control via real-time external MPC. Unlike other work on real-time MPC for quadrotors which focus on trajectory control~\cite{bangura2014real,abdolhosseini2013efficient}, ours uses a dynamics model derived fully from in-flight data that takes motors signals as direct inputs.
Effectively, our model encompasses only the actual dynamics of the system, while other implementations learn the dynamics conditioned on previously existing internal controllers.
The general nature of our model from sensors to actuators demonstrates the potential for use on robots with no previous controller --- we only use the quadrotor as the basis for comparison and do not expect it to be the limits of the MBRL system's functionality.

\subsection{Learning for Quadrotors}

Although learning-based approaches have been widely applied for trajectory control of quadrotors, implementations typically rely on sending controller outputs as setpoints to stable on-board attitude and thrust controllers. 
Iterative learning control~(ILC) approaches~\cite{schoellig2012optimization,sferrazza2017trajectory} have demonstrated robust control of quadrotor flight trajectories but require these on-board controllers for attitude setpoints. 
Learning-based model predictive control implementations, which successfully track trajectories, also wrap their control around on-board attitude controllers by directly sending Euler angle or thrust commands~\cite{bouffard2012learning,koller2018learning}. 
Gaussian process-based automatic tuning of position controller gains has been demonstrated~\cite{berkenkamp2016safe}, but only in parallel with on-board controllers tuned separately. 

Model-free reinforcement learning has been shown to generate control policies for quadrotors that out-performs linear MPC~\cite{Hwangbo2017}. 
Although similarly motivated by a desire to generate a control policy acting directly on actuator inputs, the work used an external vision system for state error correction, operated with an internal motor speed controller enabled (i.e., thrusts were commanded and not motor voltages), and generated a large fraction of its data in simulation.

Researchers of system identification for quadrotors also apply machine learning techniques. 
Bansal et al. used NN models of the Crazyflie's dynamics to plan trajectories~\cite{Bansal2016}. 
Our implementation differs by directly predicting change in attitude with on-board IMU measurements and motor voltages, rather than predicting with global, motion-capture state measurements and thrust targets for the internal PIDs.
Using Bayesian Optimization to learn a linearized quadrotor dynamics model demonstrated capabilities for tuning of an optimal control scheme~\cite{Bansal2017}. 
While this approach is data-efficient and is shown to outperform analytic models, the model learned is task-dependent. Our MBRL approach is task agnostic by only requiring a change in objective function and no new dynamics data for a new function.

\subsection{Model-based Reinforcement Learning}
Functionality of MBRL is evident in simulation for multiple tasks in low data regimes, including quadrupeds~\cite{clavera2018learning} and manipulation tasks~ \cite{kupcsik2017model}. 
Low-level MBRL control (i.e., with direct motor input signals) of an RC car has been demonstrated experimentally, but the system is of lower dimensionality and has static stability~\cite{abbeel2008apprenticeship}. 
Relatively low-level control (i.e., mostly thrust commands only passed through an internal governor before conversion to motor signals) of an autonomous helicopter has been demonstrated, but required a ground-based vision system for error correction in state estimates as well as expert demonstration for model training~\cite{abbeel2008apprenticeship}.

Properly optimized NNs trained on experimental data show test error below common analytic dynamics models for flying vehicles, but the models did not include direct actuator signals and did not include experimental validation through controller implementation~\cite{Punjani2015}. 
A model predictive path integral~(MPPI) controller using a learned NN demonstrated data-efficient trajectory control of a quadrotor, but results were only shown in simulation and required the network to be initialized with 30 minutes of demonstration data with on-board controllers~\cite{Williams2017}.

MBRL with trajectory sampling for control outperforms, in terms of samples needed for convergence, the asymptotic performance of recent model free algorithms in low dimensional tasks \cite{Chua2018}. 
Our work builds on strategies presented, with most influence derived from ``probabilistic'' NNs, to demonstrate functionality in an experimental setting --- i.e., in the presence of real-world higher order effects, variability, and time constraints.

NN-based dynamics models with MPC have functioned for experimental control of an under-actuated hexapod \cite{nagabandi2017neural}. 
The hexapod platform does not have the same requirements on frequency or control error due to its static stability, and incorporates a GPS unit for relatively low-noise state measurements.
Our work has a similar architecture, but has improvements in the network model and model predictive controller to allow substantially higher control frequencies with noisy state data. 
By demonstrating functionality without global positioning data, the procedure can be extended to more robot platforms where only internal state and actuator commands are available to create a dynamics model and control policy.

%% file: 3_groundwork.tex
\section{Experimental Setup}

\begin{figure}[t]
    \centering
    \includegraphics[width=0.7\columnwidth]{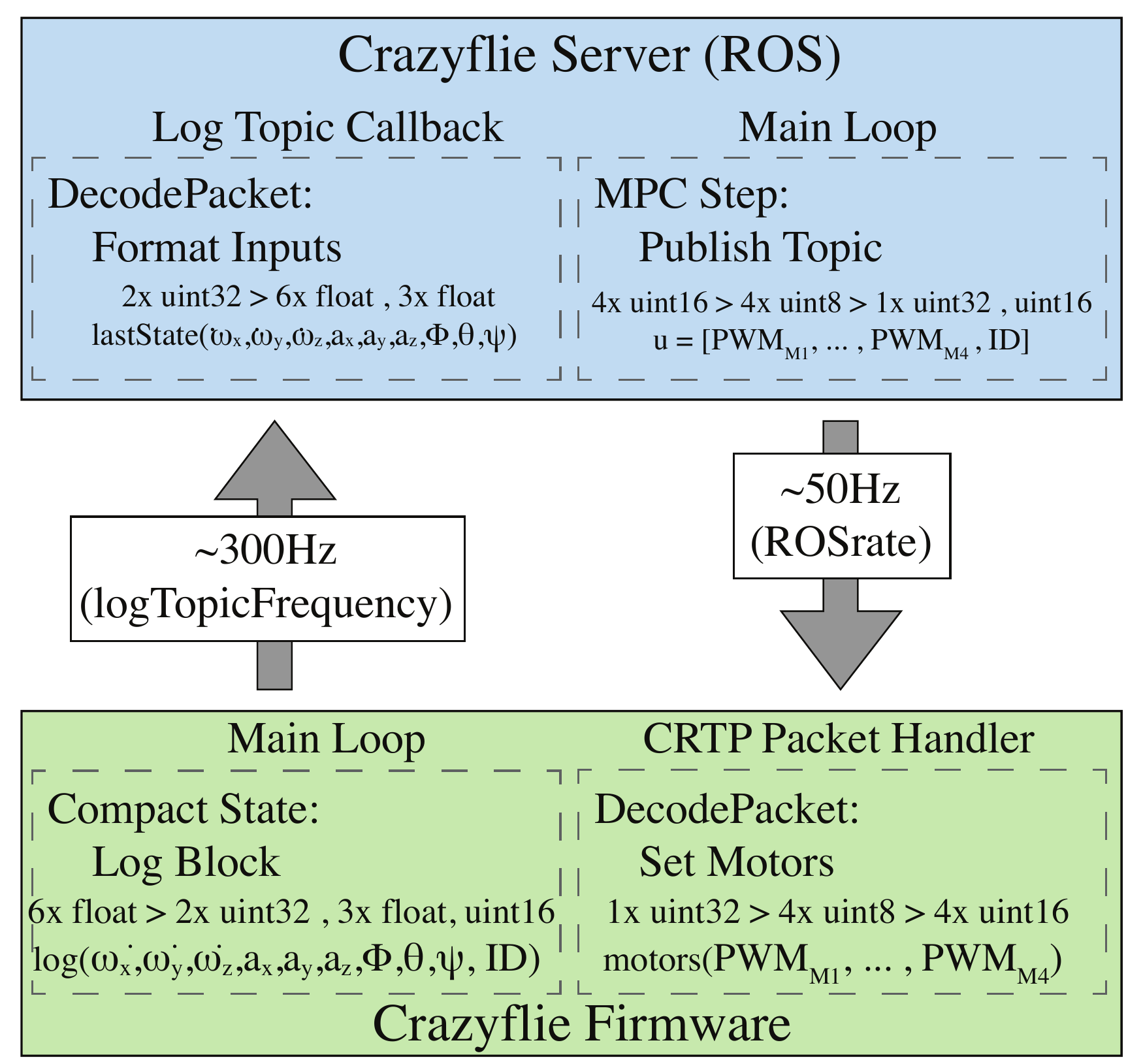}
    \caption{
    The ROS computer passes control signals and state data between the MPC node and the Crazyflie ROS server. The Crazyflie ROS server packages Tx PWM values to send and unpacks Rx compressed log data from the robot.}
    \label{fig:ROSsystem}
\end{figure}

In this paper, we use as experimental hardware platform the open-source Crazyflie 2.0 quadrotor~\cite{bitcraze2016crazyflie}.
The Crazyflie is \SI{27}{\gram} and \SI{9}{cm^2}, so the rapid system dynamics create a need for a robust controller; by default, the internal PID controller used for attitude control runs at \SI{500}{\hertz}, with Euler angle state estimation updates at \SI{1}{kHz}. 
This section specifies the ROS base-station and the firmware modifications required for external stability control of the Crazyflie.

All components we used are based on publicly available and open source projects.
We used the Crazyflie ROS interface supported here: \href{https://github.com/whoenig/crazyflie_ros}{github.com/whoenig/crazyflie\text{\_}ros}~\cite{honig2017flying}. 
This interface allows for easy modification of the radio communication and employment of the learning framework. 
Our ROS structure is simple, with a Crazyflie subscribing to PWM values generated by a controller node, which processes radio packets sent from the quadrotor in order to pass state variables to the model predictive controller (as shown in \fig{fig:ROSsystem}).
The Crazyradio PA USB radio is used to send commands from the ROS server; software settings in the included client increase the maximum data transmission bitrate up to \SI{2}{Mbps} and a Crazyflie firmware modification improves the maximum traffic rate from \SI{100}{\hertz} to \SI{400}{\hertz}.

In packaged radio transmissions from the ROS server we define actions directly as the pulse-width modulation~(PWM) signals sent to the motors.
To assign these PWM values directly to the motors we bypass the controller updates in the standard Crazyflie firmware by changing the motor power distribution whenever a CRTP Commander packet is received (see \fig{fig:ROSsystem}). 
The Crazyflie ROS package sends empty ping packets to the Crazyflie to ask for logging data in the returning acknowledgment packet; without decreasing the logging payload and rate we could not simultaneously transmit PWM commands at the desired frequency due to radio communication constraints. We created a new internal logging block of compressed IMU data and Euler angle measurements to decrease the required bitrate for logging state information, trading state measurement precision for update frequency.
Action commands and logged state data are communicated asynchronously; the ROS server control loop has a frequency set by the ROS rate command, while state data is logged based on a separate ROS topic frequency. To verify control frequency and reconstruct state action pairs during autonomous rollouts we use a round-trip packet ID system.

%% file: 4_modeltraining.tex
\section{Learning Forward Dynamics}
\label{sec:modeltraining}

\begin{figure}[t]
    \centering
    \includegraphics[width=.6\columnwidth]{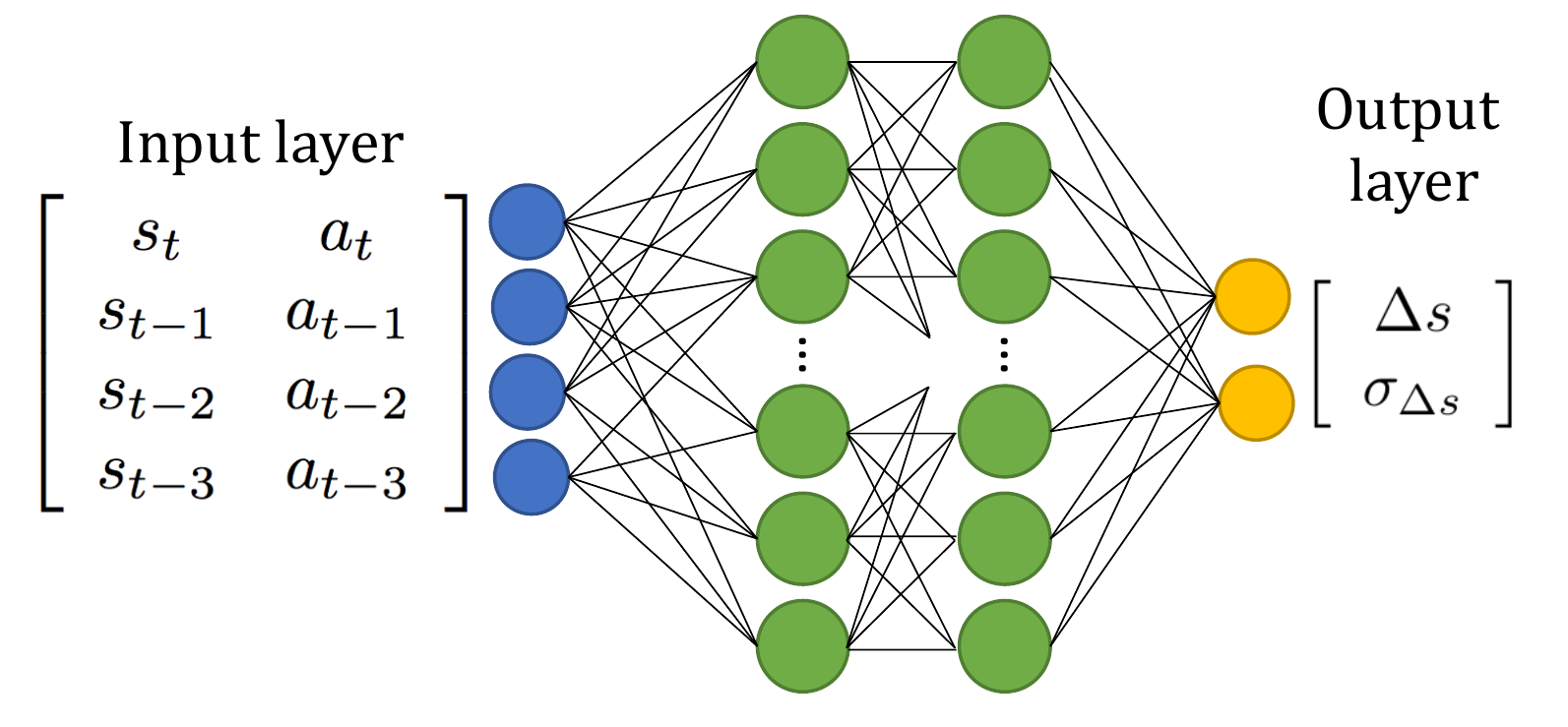}
    \caption{The NN dynamics model predicts the mean and variance of the change in state given the past 4 state-action pairs.
    We use 2 hidden layers of width 250 neurons. 
    }
    \label{fig:nn}
\end{figure}
The foundation of a controller in MBRL is a reliable forward dynamics model for predictions. 
In this paper, we refer to the current state and action as $s_t$ and $a_t$, which evolve according to the dynamics $f(s_t,a_t)$.
Generating a dynamics model for the robot often consists of training a NN to fit a parametric function $f_\theta$ to predict the next state of the robot as a discrete change in state $s_{t+1} = s_t + f_\theta(s_t,a_t)$. 
In training, using a probabilistic loss function with a penalty term on the variance of estimates, as shown in \eq{pll}, better clusters predictions for more stable predictions across multiple time-steps~\cite{Chua2018}. 
The probabilistic loss fits a Gaussian distribution to each output of the network, represented in total by a mean vector $\mu_\theta$ and a covariance matrix $\Sigma_\theta$
\begin{align}
\label{pll}
l =  \sum_{n=1}^N & [\mu_\theta (s_n, a_n) - s_{n+1}]^T \Sigma_{\theta}^{-1} (s_n, a_n) [\mu_\theta (s_n, a_n) - s_{n+1}] \nonumber \\
 & + \text{log det } \Sigma_\theta (s_n, a_n)\,. 
\end{align}
The probabilistic loss function assists model convergence and the variance penalty helps maintain stable predictions on longer time horizons. 
Our networks implemented in Pytorch train with the Adam optimizer~\cite{kingma2014adam} for 60 epochs with a learning rate of $.0005$ and a batch size of 32. 
\fig{fig:nn} summarizes the network design. 
All layers except for the output layer use the Swish activation function~\cite{ramachandran2017swish} with parameter $\beta = 1$.
The network structure was cross validated offline for prediction accuracy verses potential control frequency.
Initial validation of training parameters was done on early experiments, and the final values are held constant for each rollout in the experiments reported in \sec{sec:eval}. 
The validation set is a random subset of measured $(s_t, a_t, s_{t+1})$ tuples in the pruned data.

Additional dynamics model accuracy could be gained with systematic model verification between rollouts, but experimental variation in the current setup would limit empirical insight and a lower model loss does not guarantee improved flight time.
Our initial experiments indicate improved flight performance with forward dynamics models minimizing the mean and variance of state predictions versus models minimizing mean squared prediction error, but more experiments are needed to state clear relationships between more model parameters and flight performance.

Training a probabilistic NN to approximate the dynamics model requires pruning of logged data (e.g. dropped packets) and scaling of variables to assist model convergence. 
Our state~$s_t$ is the vector of Euler angles (yaw, pitch, and roll), linear accelerations, and angular accelerations, reading 
\begin{equation}
    s_t = \big[\dot{\omega}_x, \ \dot{\omega}_y,\ \dot{\omega}_z,\ \phi,\  \theta,\  \psi,\  \ddot{x}, \ \ddot{y}, \  \ddot{z} \big]^T.
\end{equation}

The Euler angles are from the an internal complementary filter, while the linear and angular accelerations are measured directly from the on-board MPU-9250 9-axis IMU.
In practice, for predicting across longer time horizons, modeling acceleration values as a global next state rather than a change in state increased the length of time horizon in composed predictions before the models diverged. 
While the change in Euler angle predictions are stable, the change in raw accelerations vary widely with sensor noise and cause non-physical dynamics predictions, so all the linear and angular accelerations are trained to fit the global next state. 

We combine the state data with the four PWM values, $a_t = [ m_1, m_2, m_3, m_4 ]^T $, to get the system information at time $t$. 
The NNs are cross-validated to confirm using all state data (i.e., including the relatively noisy raw measurements) improves prediction accuracy in the change in state.  

While the dynamics for a quadrotor are often represented as a linear system, for a Micro Air Vehicle (MAV) at high control frequencies motor step response and thrust asymmetry heavily impact the change in state, resulting in a heavily nonlinear dynamics model.
The step response of a Crazyflie motor RPM from PWM 0 to max or from max to 0 is on the order of \SI{250}{ms}, so our update time-step of \SI{20}{ms} is short enough for motor spin-up to contribute to learned dynamics.
To account for spin-up, we append past system information to the current state and PWMs to generate an input into the NN model that includes past time. 
From the exponential step response and with a bounded possible PWM value within $p_{eq} \pm 5000$, the motors need approximately \SI{25}{ms} to reach the desired rotor speed; when operating at \SI{50}{\hertz}, the time step between updates is \SI{20}{ms}, leading us to an appended states and PWMs history of length 4. This state action history length was validated as having the lowest test error on our data-set (lengths 1 to 10 evaluated).
This yields the final input of length 52 to our NN, $\xi$, with states and actions combined to $\xi_t = \big[s_t \ s_{t-1} \ s_{t-2} \ s_{t-3} \ a_t \ a_{t-1} \  a_{t-2} \ a_{t-3} \big]^T$.

%% file: 5_control.tex
\section{Low Level Model-Based Control}
\begin{figure}[t!]
    \centering
    \includegraphics[width=.8\columnwidth]{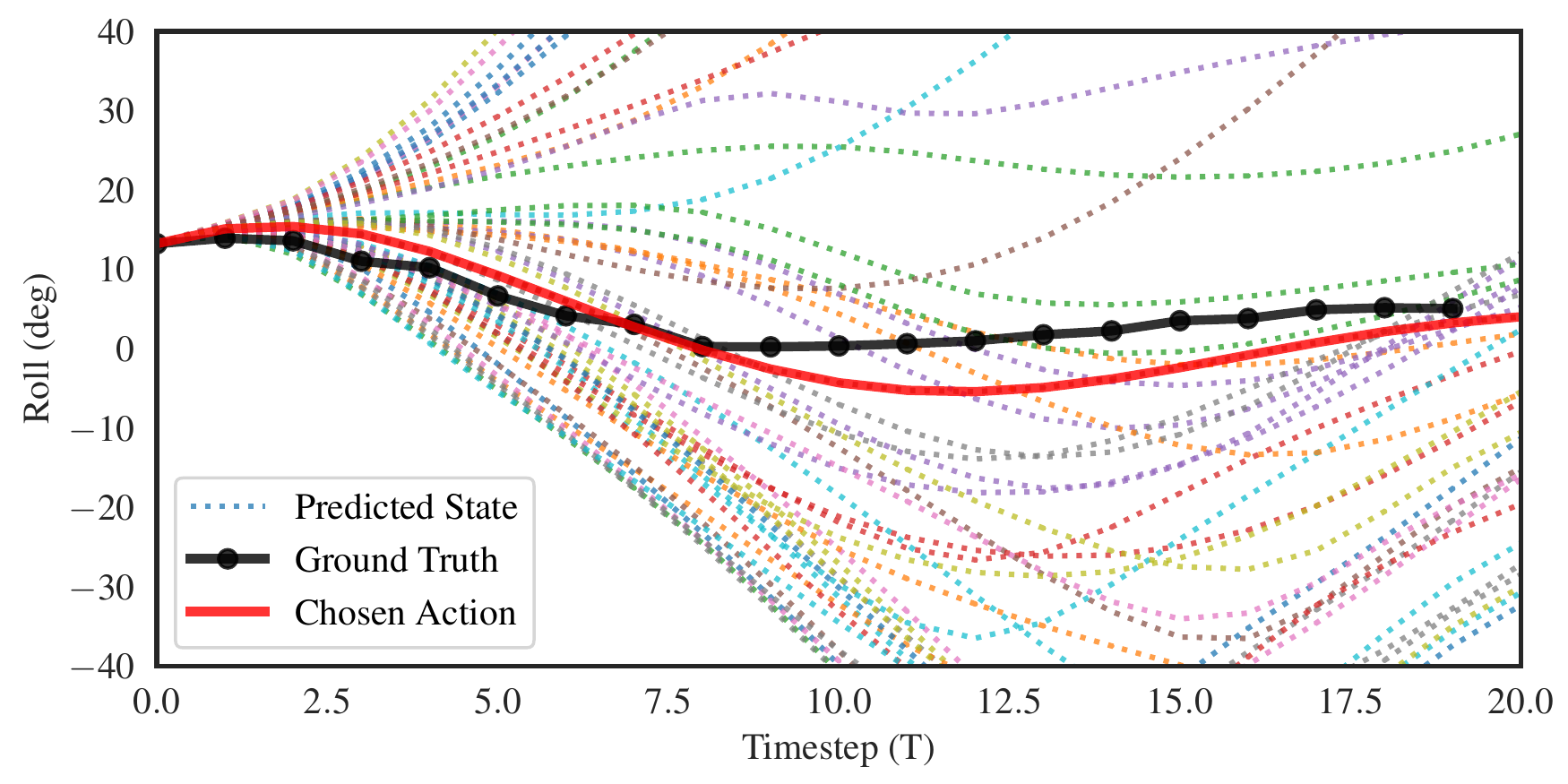}
    \caption{Predicted states for $N=50$ candidate actions with the chosen ``best action'' highlighted in red. 
    The predicted state evolution is expected to diverge from the ground truth for future $t$ because actions are re-planned at every step.}
    \label{fig:mpc}
\end{figure}

This section explains how we incorporate our learned forward dynamics model into a functional controller. 
The dynamics model is used for control by predicting the state evolution given a certain action, and the MPC provides a framework for evaluating many action candidates simultaneously.
We employ a `random shooter' MPC, where a set of $N$ randomly generated actions are simulated over a time horizon~$T$. 
The best action is decided by a user designed objective function that takes in the simulated trajectories $\hat{X}(a,s_t)$ and returns a best action, $a^*$, as visualized in \fig{fig:mpc}. 
The objective function minimizes the receding horizon cost of each state from the end of the prediction window to the current measurement.

The candidate actions, $\{a_{i} = (a_{i,1}, a_{i,2}, a_{i,3}, a_{i,4})\}_{i=1}^N$, are 4-tuples of motor PWM values centered around the stable hover-point for the Crazyflie. 
The candidate actions are constant across the prediction time horizon $T$.
For a single sample $a_{i}$, each $a_{i,j}$ is chosen from a uniform random variable on the interval $[p_{eq,j} - \sigma, p_{eq,j} + \sigma]$, where $p_{eq,j}$ is the equilibrium PWM value for motor $j$. 
The range of the uniform distribution is controlled by the tuned parameter $\sigma$; this has the effect of restricting the variety of actions the Crazyflie can take. 
For the given range of PWM values for each motor, $[p_{eq} - \sigma, p_{eq} + \sigma]$, we discretize the candidate PWM values to a step size of 256 to match the future compression into a radio packet.  
This discretization of available action choices increases the coverage of the candidate action space.
The compression of PWM resolution, while helpful for sampling and communication, represents an uncharacterized detriment to performance.

Our investigation focuses on controlled hovering, but other tasks could be commanded with a simple change to the objective function.
The objective we designed for stability seeks to minimize pitch and roll, while adding additional cost terms to Euler angle rates. 
In the cost function, $\lambda$ effects the ratio between proportional and derivative gains. 
Adding cost terms to predicted accelerations did not improve performance because of the variance of the predictions.
\begin{align}
    a^* 
    = \argmin_a \sum_{t=1}^T   \lambda (\psi_{t}^2 + \theta_{t}^2) +  \dot{\psi}_{t}^2 + \dot{\theta}_{t}^2 + \dot{\phi}_{t}^2\,.
\end{align}

Our MPC operates on a time horizon $T=12$ to leverage the predictive power of our model.
Higher control frequencies can run at a cost of prediction horizon, such as $T=9$ at \SI{75}{\hertz} or $T=6$ at \SI{100}{\hertz}. 
The computational cost is proportional to the product of model size, number of actions ($N$), and time horizon ($T$).
At high frequencies the time spanned by the dynamics model predictions shrinks because of a smaller dynamics step in prediction and by having less computation for longer $T$, limiting performance.  
At \SI{50}{\hertz}, a time horizon of 12 corresponds to a prediction of \SI{240}{ms} into the future.
Tuning the parameters of this methodology corresponds to changes in the likelihood of taking the best action, rather than modifying actuator responses, and therefore its effect on performance is less sensitive than changes to PID or standard controller parameters. 
At \SI{50}{\hertz}, the predictive power is strong, but the relatively low control frequencies increases susceptibility to disturbances in between control updates.
A system running with an Nvidia Titan Xp attains a maximum control frequency of \SI{230}{\hertz} with $N= 5000, T=1$. 
For testing we use locked frequencies of \SI{25}{\hertz} and \SI{50}{\hertz} at $N=5000, \ T=12$.

%% file: 6_eval.tex
\section{Experimental Evaluation}
\label{sec:eval}
We now describe the setting used in our experiments, the learning process of the system, and the performance summary of the control algorithm.
Videos of the flying quadrotor, and full code for controlling the Crazyflie and reproducing the experiments, are available online at \url{https://sites.google.com/berkeley.edu/mbrl-quadrotor/}

\subsection{Experimental Setting}
The performance of our controller is measured by the average flight length over each roll-out. 
Failure is often due to drift induced collisions, or, as in many earlier roll-outs, when flights reach a pitch or roll angle over \SI{40}{\degree}. 
In both cases, an emergency stop command is sent to the motors to minimize damage. 
Additionally, the simple on-board state estimator shows heavy inconsistencies on the Euler angles following a rapid throttle ramping, which is a potential limiting factor on the length of controlled flight. 
Notably, a quadrotor with internal PIDs enabled will still fail regularly due to drift on the same time frame as our controller; it is only with external inputs that the internal controllers will obtain substantially longer flights.
The drift showcases the challenge of using attitude controllers to mitigate an offset in velocity.

\begin{figure}[t]
    \centering
    \includegraphics[width=.8\columnwidth]{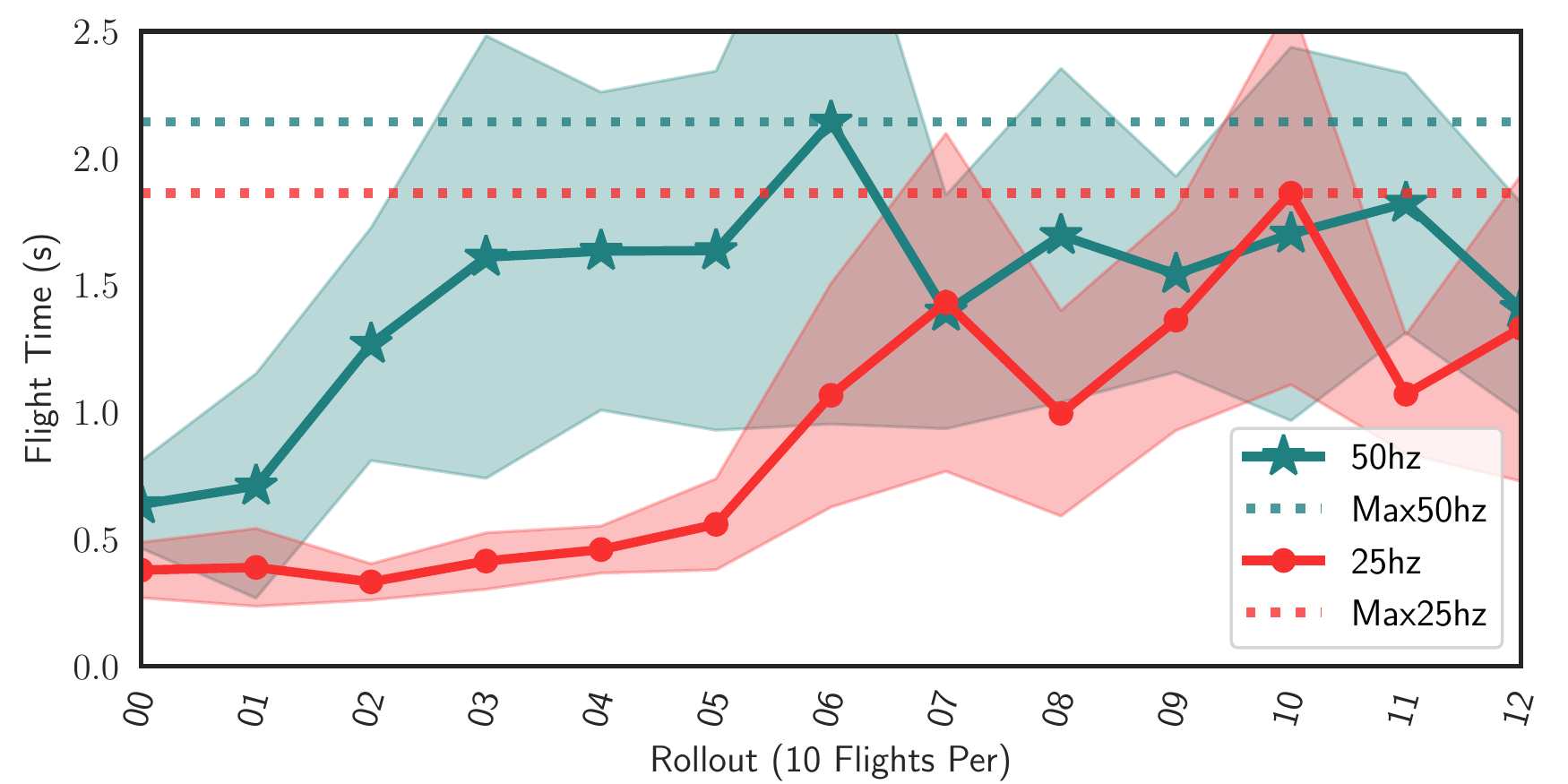}
    \caption{Mean and standard deviation of the 10 flights during each rollout learning at \SI{25}{\hertz} and \SI{50}{\hertz}. The \SI{50}{\hertz} shows a slight edge on final performance, but a much quicker learning ability per flight by having more action changes during control.}
    \label{fig:learn}
\end{figure}

\begin{figure*}[t]
    \centering
    
    \includegraphics[width=.80\textwidth]{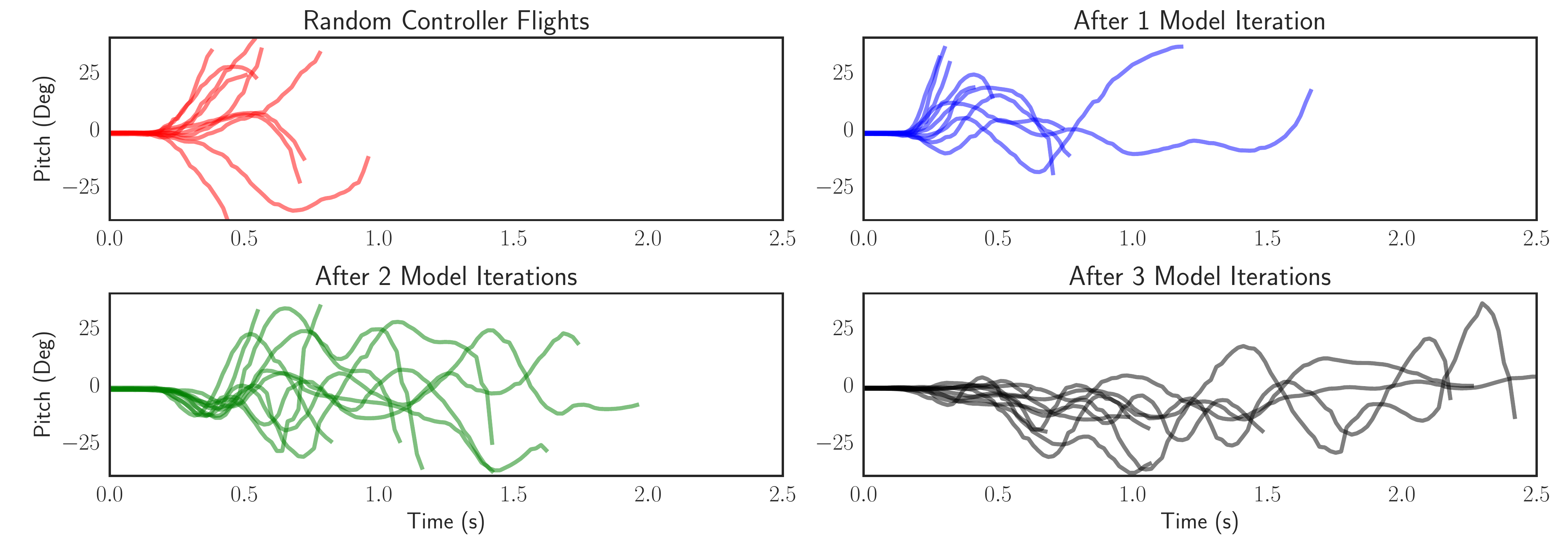}
    \caption{The pitch over time for each flight in the first four roll-outs of learning at \SI{50}{\hertz}, showing the rapid increase in control ability on limited data.
    The random and first controlled roll-out show little ability, but roll-out 3 is already flying for $>2$~seconds.
    }
    \label{fig:roll}
\end{figure*}

\subsection{Learning Process}

The learning process follows the RL framework of collecting data and iteratively updating the policy. 
We trained an initial model $f_0$ on 124 and 394 points of dynamics data at \SI{25}{\hertz} and \SI{50}{\hertz}, respectively, from the Crazyflie being flown by a random action controller.
Starting with this initial model as the MPC plant, the Crazyflie undertakes a series of autonomous flights from the ground with a \SI{250}{ms} ramp up, open-loop takeoff followed by on-policy control while logging data via radio.
Each roll-out is a series of 10 flights, which causes large variances in flight time. 
The initial roll-outs have less control authority and inherently explore more extreme attitude orientations (often during crashes), which is valuable to future iterations that wish to recover from higher pitch and/or roll.
The random and first three controlled roll-outs at \SI{50}{\hertz} are plotted in \fig{fig:roll} to show the rapid improvement of performance with little training data.

The full learning curves are shown in \fig{fig:learn}.
At both \SI{25}{\hertz} and \SI{50}{\hertz} the rate of flight improvement reaches its maximum once there is 1,000 trainable points for the dynamics model, which takes longer to collect at the lower control frequency.
The improvement is after roll-out 1 at 50Hz and roll-out 5 at 25Hz.
The longest individual flights at both control frequencies is over \SI{5}{\second}. 
The final models at \SI{25}{\hertz} and \SI{50}{\hertz} are trained on 2,608 and 9,655 points respectively, but peak performance is earlier due to dynamics model convergence and hardware lifetime limitations.

\begin{figure}[t]
    \centering
    \includegraphics[width=.88\columnwidth]{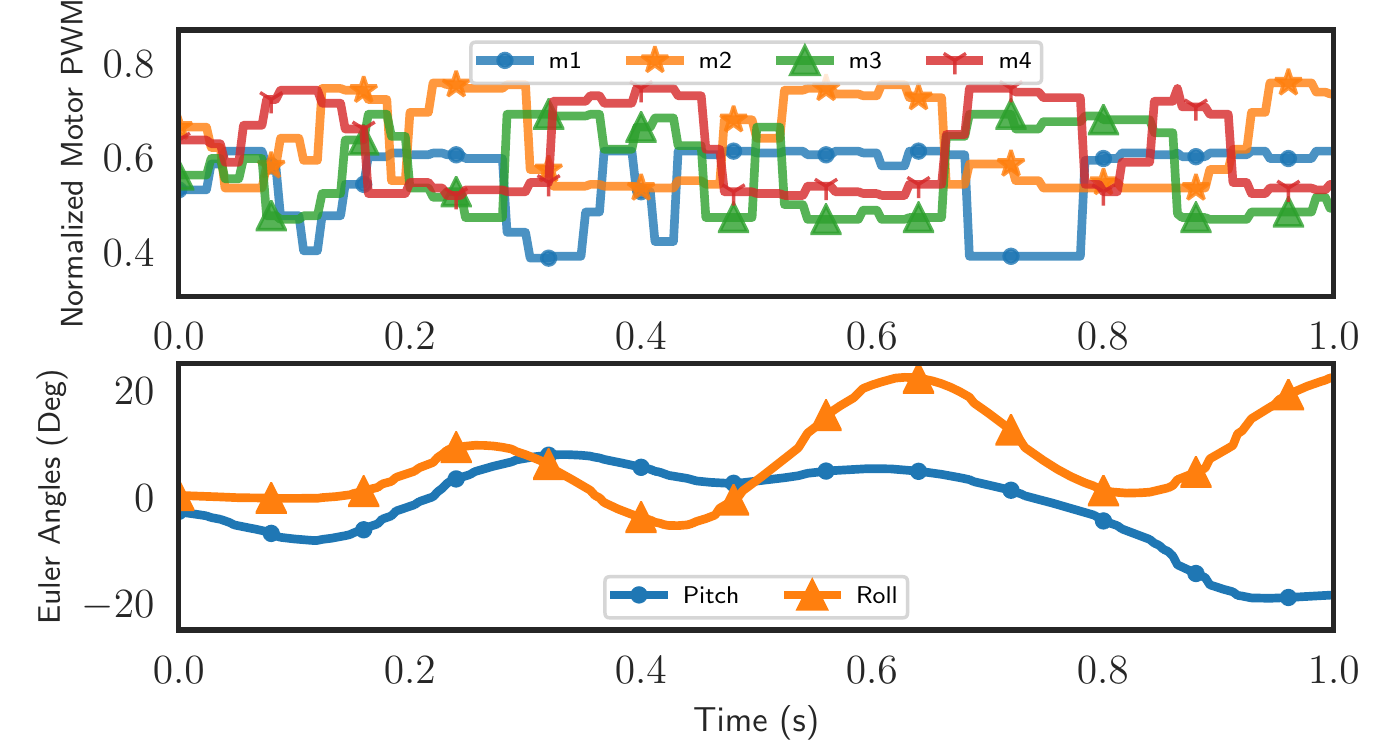}
    \caption{The performance of the \SI{50}{\hertz} controller.  (\textit{Above}) The controlled PWM values over time, which visibly change in response to angle oscillations. (\textit{Below})  Pitch and roll.}
    \label{fig:eval}
\end{figure}

\subsection{Performance Summary}
\begin{figure*}[t]
    \centering
    \includegraphics[width=.82\textwidth]{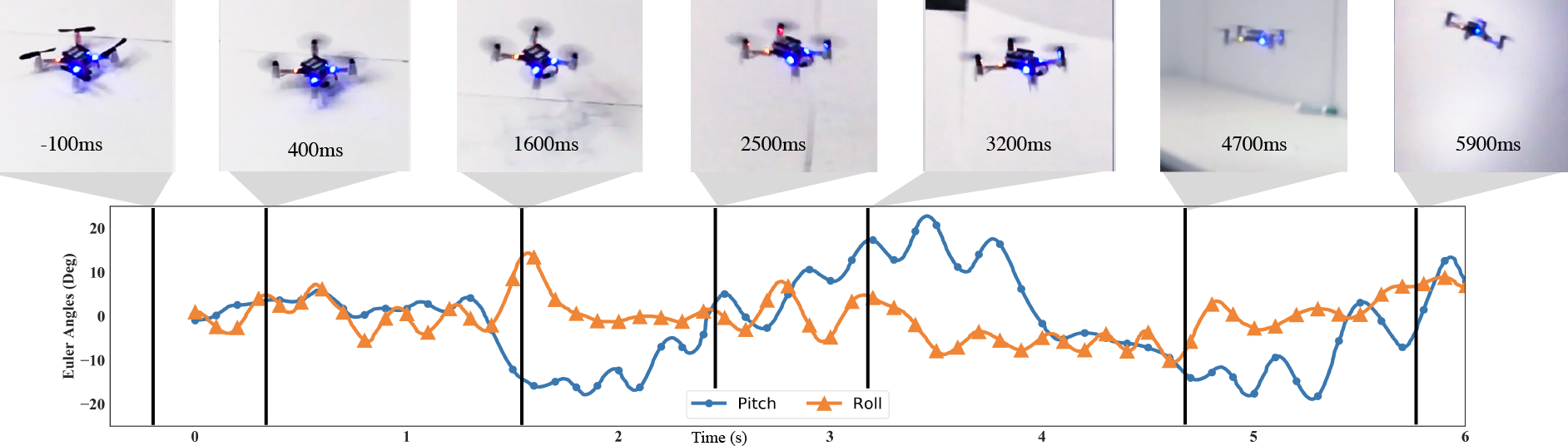}
    \caption[Caption of Photos]{A full flight of Euler angle state data with frames of the corresponding video. This flight would have continued longer if not for drifting into the wall. The relation between physical orientation and pitch and roll is visible in the frames. The full video is online on the accompanying website.}
    \label{fig:photos}
\end{figure*}

This controller demonstrates the ability to hover, following a ``clean'' open-loop takeoff, for multiple seconds (an example is shown in \fig{fig:photos}). 
At both \SI{25}{\hertz} and \SI{50}{\hertz}, once reaching maximum performance in the 12 roll-outs, about $30\%$ of flights fail to drift.
The failures due to drift indicate the full potential of the MBRL solution to low-level quadrotor control.
An example of a test flight segment is shown in \fig{fig:eval}, where the control response to pitch and roll error is visible.

The basis of comparison, typical quadrotor controllers, achieve better performance, but with higher control frequencies and engineering design iterations leveraging system dynamics knowledge.
With the continued improvement of computational power, the performance of this method should be re-characterized as potential control frequencies approach that of PID controllers.
Beyond comparison to PID controllers with low computational footprints, the results warrant exploration of MBRL for new dynamical systems with or when varying goals need to be built into low level control.
In less than 10 minutes of clock time, and only 3 minutes of training data, we present comparable, but limited, performance that is encouraging for future abilities to match and surpass basic controllers.
Moving the balance of this work further towards domain specific control would likely improve performance, but the broad potential for applications to more and different robotic platforms compels exciting future use of MBRL.

%% file: 7_discussion.tex
\section{Discussion and Limitations}
\label{disuccsion}
The system has multiple factors contributing to short length and high variance of flights. 
First, the PWM equilibrium values of the motors shift by over $10\%$ following a collision, causing the true dynamics model to shift over time. 
This problem is partially mitigated by replacing the components of the Crazyflie, but any change of hardware causes dynamics model mismatch and the challenge persists.
Additionally, the internal state estimator does not track extreme changes in Euler angles accurately.
We believe that overcoming the system-level and dynamical limitations of controlling the Crazyflie in this manner showcases the expressive power of MBRL.

Improvements to the peak performance will come by identifying causes of the performance plateau.
Elements to investigate include the data-limited slow down in improvement of the dynamics model accuracy, the different collected data distributions at each roll-out, the stochasticity of NN training, and the stochasticity at running time with MPC.

Beyond improving performance, computational burden and safety hinder the applicability of MBRL with MPC to more systems.
The current method requires a GPU-enabled base-station, but the computational efficiency could be improved with intelligent action sampling methods or by combining model free techniques, such as learning a deterministic action policy based on the learned dynamics model. 
We are exploring methods to generate NN control policies, such as an imitative-MPC network or a model-free variant, on the dynamics model that could reduce computation by over 1000x by only evaluating a NN once per state measurement.
In order to enhance safety, we are interested in defining safety constraints within the model predictive controller, rather than just a safety kill-switch in firmware, opening the door to fully autonomous learned control from start to finish.

%% file: 8_conclusion.tex
\section{Conclusions and Future Work}
This work is an exploration of the capabilities of model-based reinforcement learning for low-level control of an \textit{a priori} unknown dynamic system.
The results, with the added challenges of the static instability and fast dynamics of the Crazyflie, show the capabilities and future potential of MBRL.
We detail the firmware modifications, system design, and model learning considerations required to enable the use of a MBRL-based MPC system for quadrotor control over radio.
We removed all robot-specific transforms and higher level commands to only design the controller on top of a learned dynamics model to accomplish a simple task.
The controller shows the capability to hover for multiple seconds at a time with less than 3 minutes of collected data -- approximately half of the full battery life flight time of a Crazyflie. 
With learned flight in only minutes of testing, this brand of system-agnostic MBRL is an exciting solution not only due to its generalizability, but also due to its learning speed. 

In parallel with addressing the limitations outlined in \sec{disuccsion}, the quadrotor results warrant investigation into low level control of other robots.
The emergent area of microrobotics combines the issues of under-characterized dynamics, weak or non-existent controllers, ``fast'' dynamics and therefore instabilities, and high cost-to-test~\cite{drew2018toward,contreras2017first}, so it is a strong candidate for MBRL experiments.

%% file: _appendices.tex
\clearpage
\newpage


\section{Appendix}

\csvstyle{myTableStyle}{tabular=|l||c|c|c|c|c|,
    table head=\hline  R & Mean Time (ms) & Std. Dev.  Time  &  Collected Points & Total Trained Points & RMS pitch \& roll \\\hline\hline,
    late after line=\\\hline,
    head to column names}

\subsection{Battery Voltage Context}
\label{Battery Voltage}

The Crazyflie has a short battery voltage of about 7 minutes of flight time and operation depends heavily on battery voltage, with it becoming uncontrollable on low voltages when operating on our controllers or the built in nested PID controllers. 
In this experiment, we study the influence of the battery voltage to the dynamics of the Crazyflie, to understand if there is a time-varying drift that need to be compensated.
We investigate this hypothesis by logging battery voltage and adding it to the state passed to the neural network during model training to improve prediction accuracy. 

\begin{figure*}[b!]
    \centering
    \includegraphics[width=\textwidth]{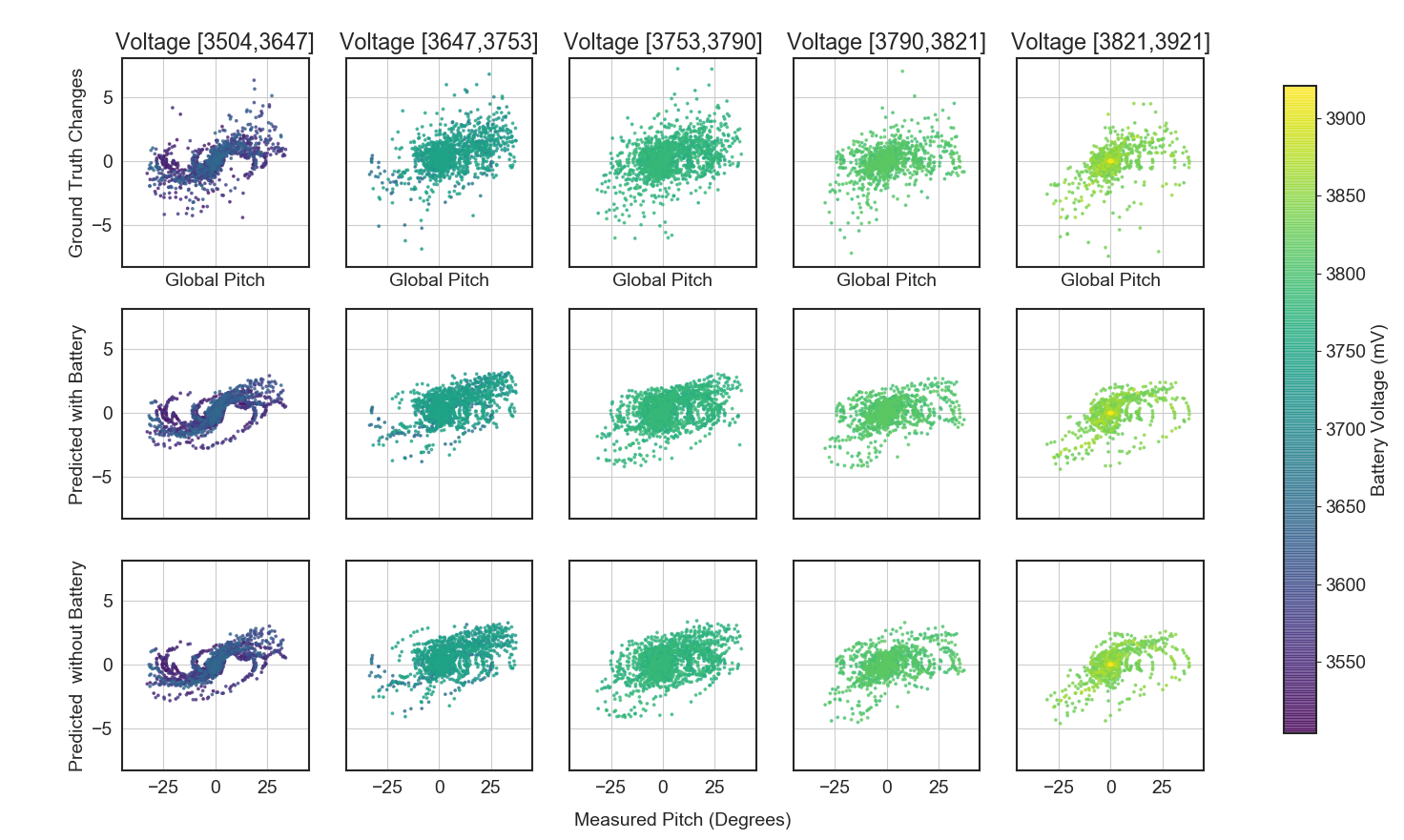}
    \caption{Demonstration of the effect of battery voltage on state predictions. The top row is the ground truth one step pitch changes, the middle is the predictions through a model trained with battery voltage included in the state, and the bottom is predictions without battery voltage included in the state. Both of the predictions show tighter grouping from the variance term on the probabilistic loss function, but there is an extremely low difference between the predictions with and the predictions without battery. The lack of difference in predictions indicates the battery voltage is latent to other variables passed into the network.}
    \label{fig:vbat}
\end{figure*}

When operating the Crazyflie at control frequencies of greater than \SI{100}{\hertz}, the state dynamics become clearly biased at battery voltages less than 3,650 mV. 
The biases are present at lower frequencies, but less pronounced.
The biased state dynamics can be seen in \fig{fig:vbat}, but the predictions do not improve when passing the battery voltage into the neural network dynamics model at any battery level.
The RMS error delta between a model trained with and without battery voltage is less than 1\%, indicating that the battery voltage is nearly completely captured in other variables passed to the network. 

\begin{figure}[t]
    \centering
    \includegraphics[width=\columnwidth]{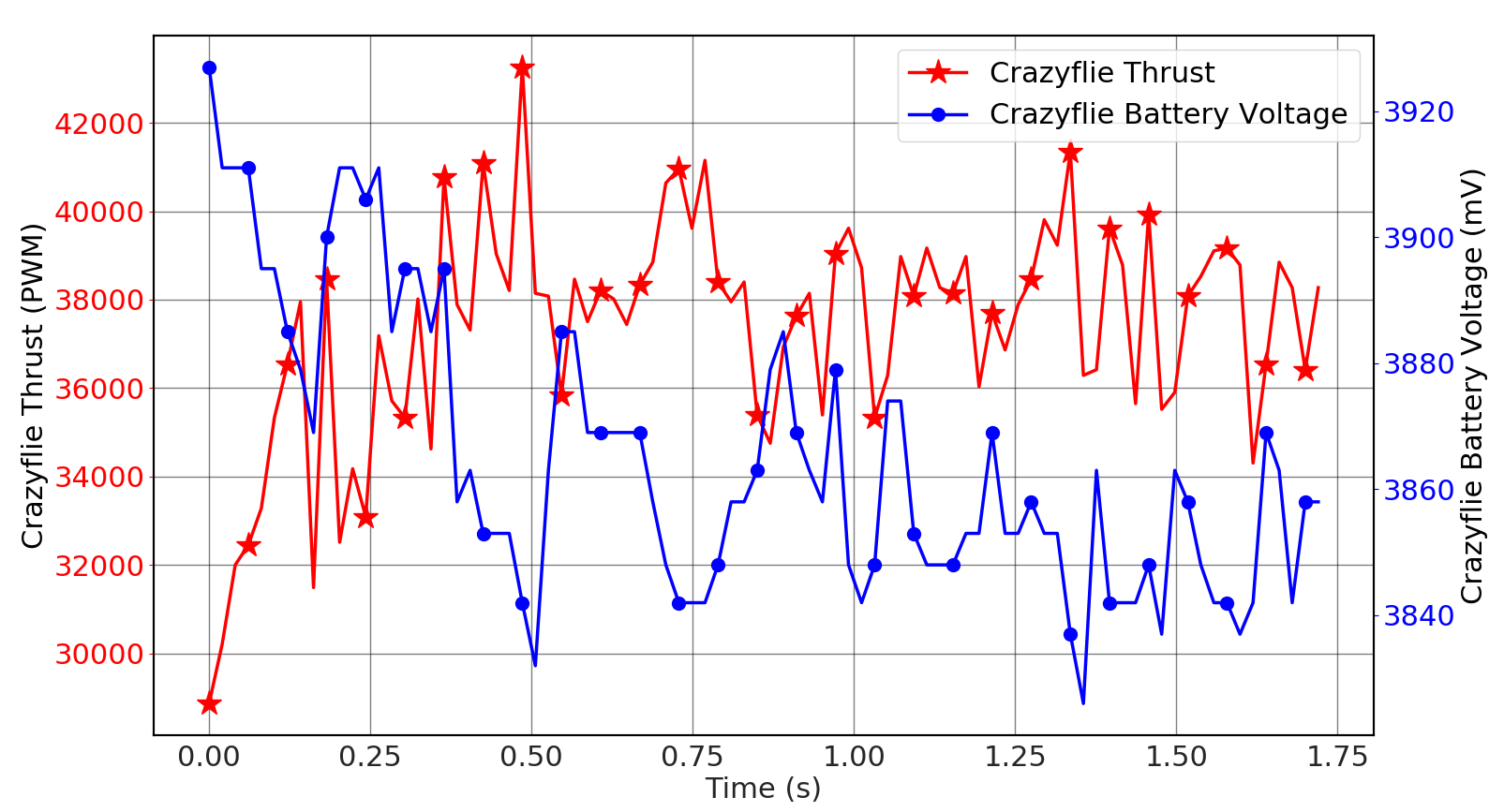}
    \caption{The logged battery voltage and mean PWM of the 4 motors across a flight. There is a clear inverse relationship between the logged battery voltage and the current thrust.}
    \label{fig:batthrust}
\end{figure}

A potential explanation for the lack of model improvement with logged battery voltage is that the current battery reading is latent in other variables past into the network, and the natural charge based variations in data are not dominant. 
The logged data shows a clear inverse relationship between battery voltage and current Crazyflie thrust, shown in \fig{fig:batthrust}. 
The impedance of the motors changes depending on the rotor speed and drive.
This battery and thrust relationship is less likely to be apparent on quadrotors with separate motor voltage controllers, where the impedance of the motors changing with revolutions per minute would be compensated for.

\subsection{Crazyflie Lifespan}
\label{Lifespan}

\begin{figure*}[t]
    \centering
    \includegraphics[width=\textwidth]{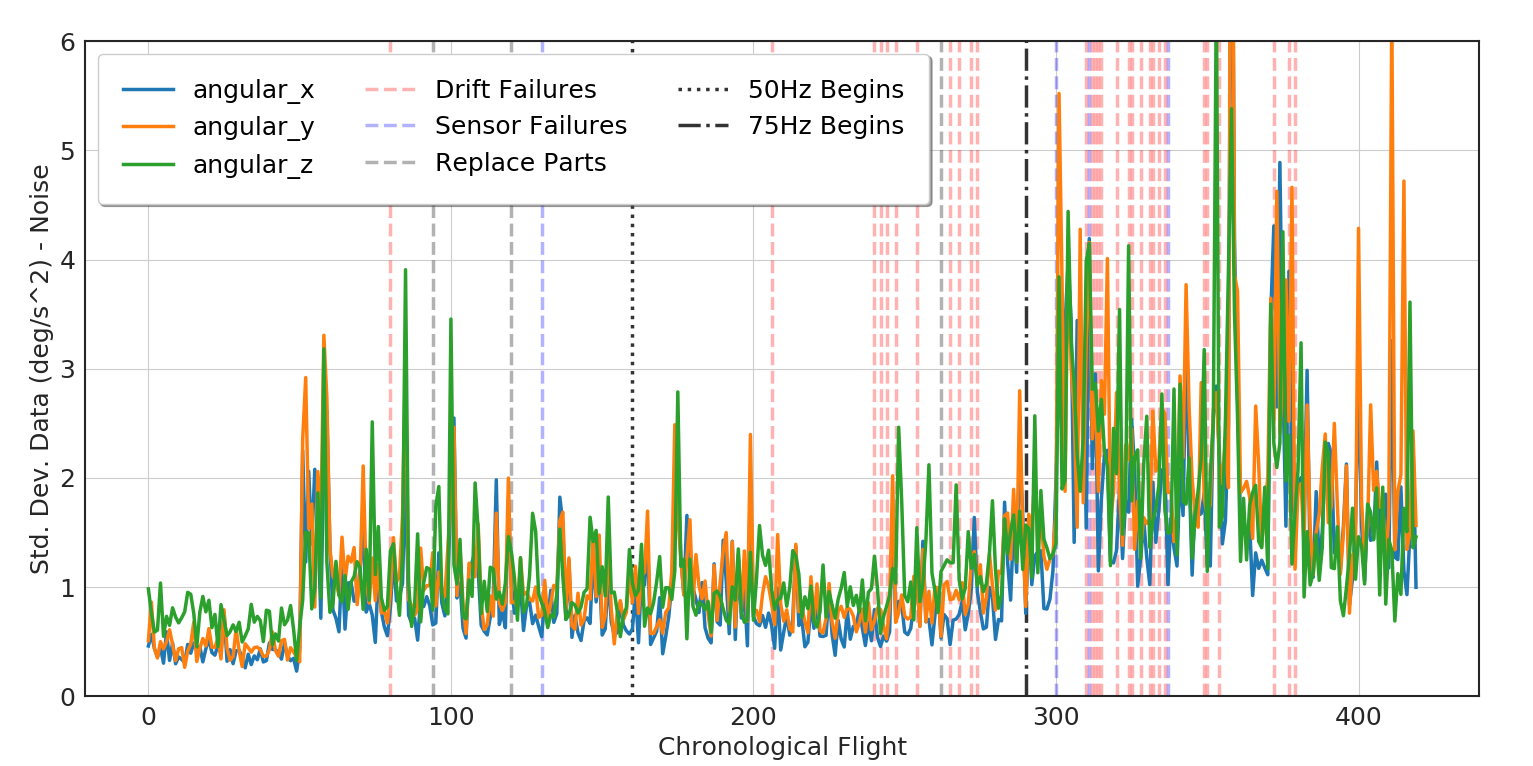}
    \caption{The sensor noise on the 3 angular accelerations measured by the gyroscope of the MPU9250 of the Crazyflie before the robot takes off. The black vertical lines separate the rollouts at \SI{25}{\hertz}, \SI{50}{\hertz} and \SI{75}{\hertz} from left to right. The vertical lines indicate changes in hardware and collisions that would change the dynamics or state of the robot. The sensors clearly are subject to increasing noise over lifespan.}
    \label{fig:life}
\end{figure*}

Extended periods of testing on individuals quadrotors demonstrated a finite lifetime. 
After many flights, performance would dip inexplicably. 
This is due to a combination of motor damage and or sensor decay. 
Motor damage causes a measureable change in the equilibrium PWMs for a given quadrotor.
\fig{fig:life} shows the change in the noise on the gyroscope before takeoff for all of the flights taken by the quadcopter used to collect data for this publication.
The left two sections includes the data included in \ref{sec:eval}, but the data taken at a control frequency of \SI{75}{\hertz} was abandoned due to inconsistent performance.
Some initial flights at \SI{75}{\hertz} were extremely promising, but after a series of collisions via drift the quadrotor would not take off cleanly.
Future work should investigate methods of mitigating the effect of sensor drift, potentially by conditioning the dynamics model on a sensor noise measurement or enforcing more safety constraints on flight.

\subsection{Frequency Dependent Learning}
\label{frequency}
\begin{figure}[h]
    \centering
    \includegraphics[width=\columnwidth]{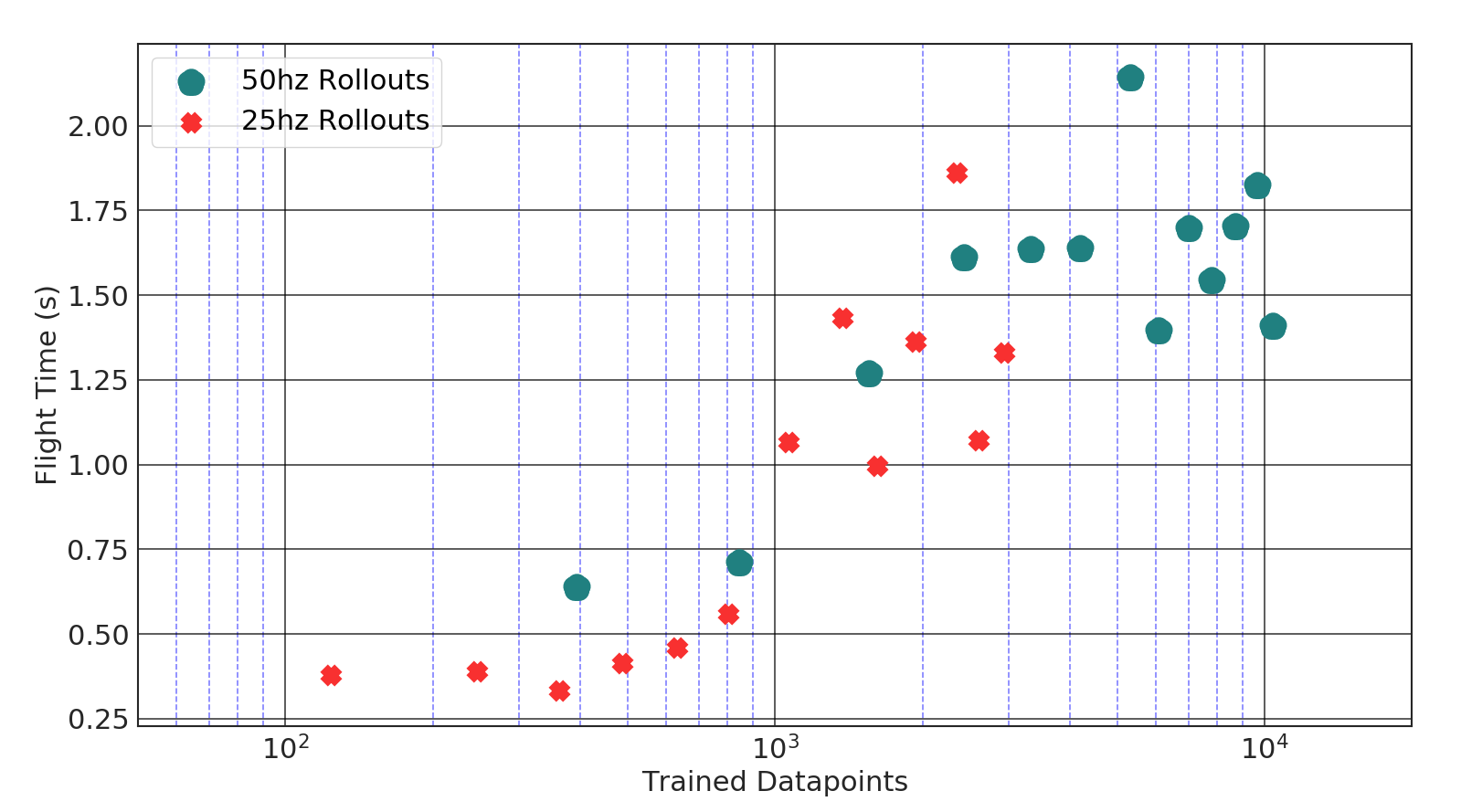}
    \caption{Mean flight time of each rollout plotted verses the logarithm of the number of available points at train time for each model. The higher control frequency allows the controller to learn faster on wall time, but the plot indicates that there is not a notable difference between control ability when the number of trained points are equal. There is a continuing upward trend of flight time verses training points, but it is difficult to obtain more data in experiment.}
    \label{fig:pts}
\end{figure}

There is a trend between learned performance at both frequencies and the number of trained points for the model, as shown in \fig{fig:pts}.
The continuing upward trend between logarithmic points and flight time indicates further data collection could enhance flight performance, but is unrealistic without further progress on safe learning with the Crazyflie.
Potential future applications could leverage a combination of our results with bootstrapping data to continue to improve performance without the difficulties of logging large amoutns of experimental data on an individual robot.

